\documentclass[letterpaper, 10 pt, conference]{ieeeconf} 

\IEEEoverridecommandlockouts
\usepackage{multicol}
\usepackage{graphicx}
\usepackage{amsmath,amsfonts,amssymb}
\usepackage[bookmarks=true]{hyperref}
\usepackage{multirow}
\usepackage{booktabs}
\usepackage{subfig}
\usepackage{tablefootnote}
\usepackage[flushleft]{threeparttable}
\usepackage{lipsum}

\begin{document}

\title{\LARGE \bf Interventional Behavior Prediction: Avoiding Overly Confident Anticipation in Interactive Prediction}

\author{Chen Tang, Wei Zhan, Masayoshi Tomizuka
\thanks{C. Tang, W. Zhan, and M. Tomizuka, are with the Department of Mechanical Engineering, University of California, Berkeley, CA 94720, USA {\tt\small 	\{chen\_tang, wzhan, tomizuka\}@berkeley.edu}.}%
\thanks{This work is supported by Denso International America, Inc.}
}



%

\maketitle

\begin{abstract}
Conditional behavior prediction (CBP) builds up the foundation for a coherent interactive prediction and planning framework that can enable more efficient and less conservative maneuvers in interactive scenarios. In CBP task, we train a prediction model approximating the posterior distribution of target agents' future trajectories conditioned on the future trajectory of an assigned ego agent. However, we argue that CBP may provide overly confident anticipation on how the autonomous agent may influence the target agents' behavior. Consequently, it is risky for the planner to query a CBP model. Instead, we should treat the planned trajectory as an intervention and let the model learn the trajectory distribution under intervention. We refer to it as the interventional behavior prediction (IBP) task. Moreover, to properly evaluate an IBP model with offline datasets, we propose a Shapley-value-based metric to verify if the prediction model satisfies the inherent temporal independence of an interventional distribution. We show that the proposed metric can effectively identify a CBP model violating the temporal independence, which plays an important role when establishing IBP benchmarks.  
\end{abstract}

\begin{figure}[t]
    \centering
    \includegraphics[width=2.6in]{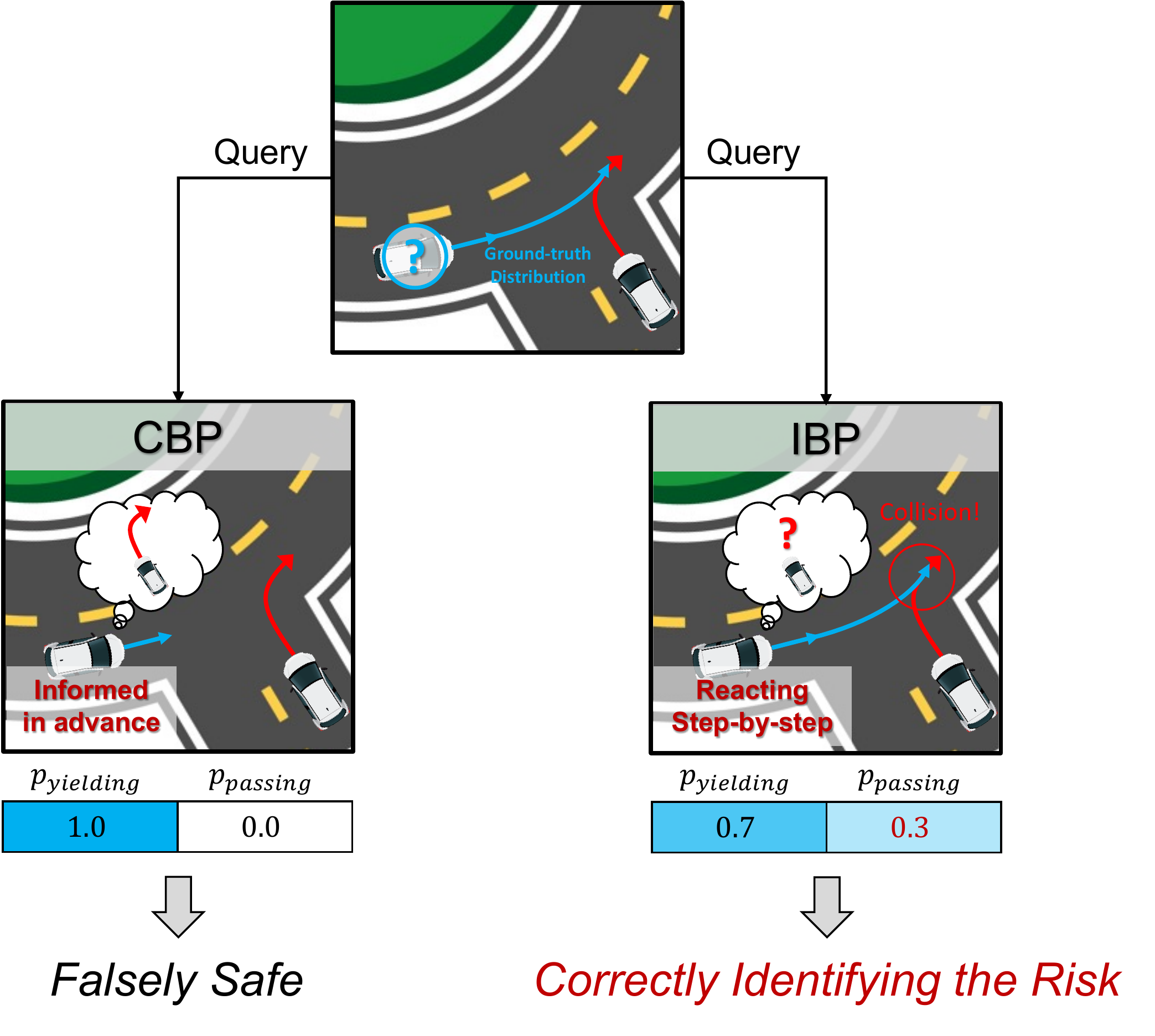}
    \caption{An illustration of the difference between CBP and IBP. The robot car plans to enter the roundabout aggressively and force the human car in the roundabout to yield. It queries a prediction model for whether the human car will yield or pass. The CBP model predicts the posterior distribution of human car's behavior conditioned on the plan. Intuitively, it models the driver behavior when he knows the robot car's plan in advance. Therefore, the human car will always be predicted to yield to the robot car under CBP. In fact, the human car will only react to the robot car's action at each timestep. Consequently, the human car may attempt to pass first as it has the right-of-way, which may lead to collision. The IBP model is able to detect the safety risk.}
    \label{fig:illustration}
    \vspace{-0.1in}
\end{figure}

\section{Introduction}
Behavior prediction is crucial for autonomous systems interacting with humans, such as autonomous vehicles. Most existing works focus on a \emph{passive} prediction scheme \cite{gu2021densetnt, tang2021exploring}, where the target agents' future trajectories are predicted given the historical trajectories of themselves and other surrounding agents. When using such a prediction model, downstream decision-making modules then determine the autonomous agent's action according to the predicted trajectories in a passive manner. To ensure safety under various predicted trajectories of others, the ego agent has to be overly conservative with inefficient maneuvers, especially in highly interactive scenarios. It is because passive prediction models ignore the fact that the autonomous agent's future actions can influence other agents' behavior. To this end, researchers started to investigate a more coherent interactive prediction and planning framework which relies on predicting the surrounding agents' future trajectories conditioned on the ego agent's future actions \cite{schmerling2018multimodal, tang2019multiple, rhinehart2019precog, khandelwal2020if, salzmann2020trajectron++, song2020pip, liu2021deep, tolstaya2021identifying}. Under such frameworks, the autonomous agents can reason over potential actions while considering their influence on surrounding agents. It can then induce more efficient and less conservative maneuvers in interactive scenes. Some of these prior works merely demonstrated that their models are able to support conditional prediction from the perspective of architecture \cite{tang2019multiple, khandelwal2020if}. Another line of works focused on the closed-loop performance, which relies on a simulating environment \cite{schmerling2018multimodal, rhinehart2019precog, liu2021deep}. 

More interestingly, some existing works formulated an alternative prediction task to evaluate the prediction module in a self-contained way \cite{song2020pip, salzmann2020trajectron++, tolstaya2021identifying}. We follow \cite{tolstaya2021identifying} to refer to this task as \emph{conditional behavior prediction} (CBP). In the CBP task, the future trajectories of the target agents are predicted conditioned on the ground-truth future trajectory of an assigned ego agent. Standard prediction metrics are adopted to quantify the performance. It allows us to leverage large-scale naturalistic traffic datasets to develop and validate a conditional prediction model before closed-loop testings. In those works, a model that can achieve the smallest prediction error after granted the additional future information of the ego agent is considered the best. However, we can only evaluate the prediction accuracy given the actual future trajectory of the ego agent with a static offline dataset. It is impossible to quantify the model performance when it is queried by an arbitrary plan of the ego agent. Therefore, we should be careful when interpreting the evaluation results. 

In particular, we argue that it is risky to train and evaluate the model for \emph{conditional inference}. In the current CBP task, the prediction model essentially learns the posterior distribution of future trajectories conditioned on the future trajectory of the ego agent. In this way, the ego agent's future trajectory is treated as an \emph{observation}. Since the actual ego agents in the offline dataset make decisions according to the states of the surrounding agents, the surrounding agents under CBP are implicitly assumed to get additional hints on the future behavior of the ego agents. With such an unrealistic assumption, it is natural to consider the CBP model with the lowest prediction error as the best option for the CBP task. However, the surrounding agents in the real world are not informed of the planned trajectories of the ego agents. Consequently, as illustrated in Fig. \ref{fig:illustration}, there will be a discrepancy between: 1) what an autonomous agent is informed by querying a CBP model with a potential plan; and 2) how the others will actually react if the agent executes the plan. As we will show later, this discrepancy may lead to overly confident anticipation on the ego agent's influence on the surroundings, resulting in potential safety hazards. 

This discrepancy is formally captured in the theory of causality \cite{reason:Pearl09a} by the difference between \emph{observation} and \emph{intervention}. With an intervention to a set of random variables, we enforce the value of a random variable without treating it as the consequence of other random variables. The resulting distribution of the remaining random variables under the intervention is consistent with what will actually happen if we have the privilege to manipulate the target random variable as desired. Consequently, we argue that we should build the prediction model to approximate the future trajectory distribution under the intervention of enforcing the ego agent's future trajectory. We refer to this new task as the \emph{interventional behavior prediction} (IBP) task. 
In IBP, we still want to train and evaluate the model with an offline dataset. The task setting is essentially the same as CBP, except for learning an interventional distribution instead of a conditional one. The remaining issue is then how to properly evaluate an IBP model with an offline dataset. Without knowing the ground-truth distribution under intervention, we can only compare the model's output against the ground-truth future trajectories for evaluation. However, such evaluation metrics are naturally biased toward a CBP model. The dataset is collected without intervention. The ego agent in the dataset follows an internal reactive policy. Therefore, the distribution of ground-truth labels given the same input essentially follows a conditional distribution. As a result, a CBP model will always outperform an IBP model if prediction accuracy is the only evaluation metric with an offline dataset. 

To this end, we propose to verify the inherent temporal independence of a prediction model before comparing the prediction performance to ensure a proper evaluation for the IBP task. Under the interventional distribution, the predicted states of the target agents at earlier timesteps should be independent from the ego agent's states at latter timesteps. We propose a Shapley-value-based \cite{RM-670-PR, lundberg2017unified, makansi2021you} metric to verify if the model obeys this temporal independence. We show that we can effectively identify a model violating the expected temporal independence with the proposed metric, and therefore ensure a valid evaluation method. 

The rest of the paper is organized as follows: 1) In Sec. \ref{sec:toy-example}, we explain the difference between CBP and IBP and demonstrate the risk of using CBP with a motivating toy example; 2) In Sec. \ref{sec:shapley}, we formulate the Shapley-value-based metric we propose to verify the temporal independence and quantify the impact of doing CBP; 3) In Sec. \ref{sec:exp}, we study a CBP model with the proposed metric to demonstrate that such a CBP model indeed violates the temporal independence. Moreover, it results in misleading evaluation result without using the proposed metric. 4) In Sec. \ref{sec:discuss}, we wrap up the paper with a discussion on insights for future model design and IBP benchmarks. 

\section{A Motivating Example} \label{sec:toy-example}

\begin{figure}[t]
    \centering
    \includegraphics[width=1.5in]{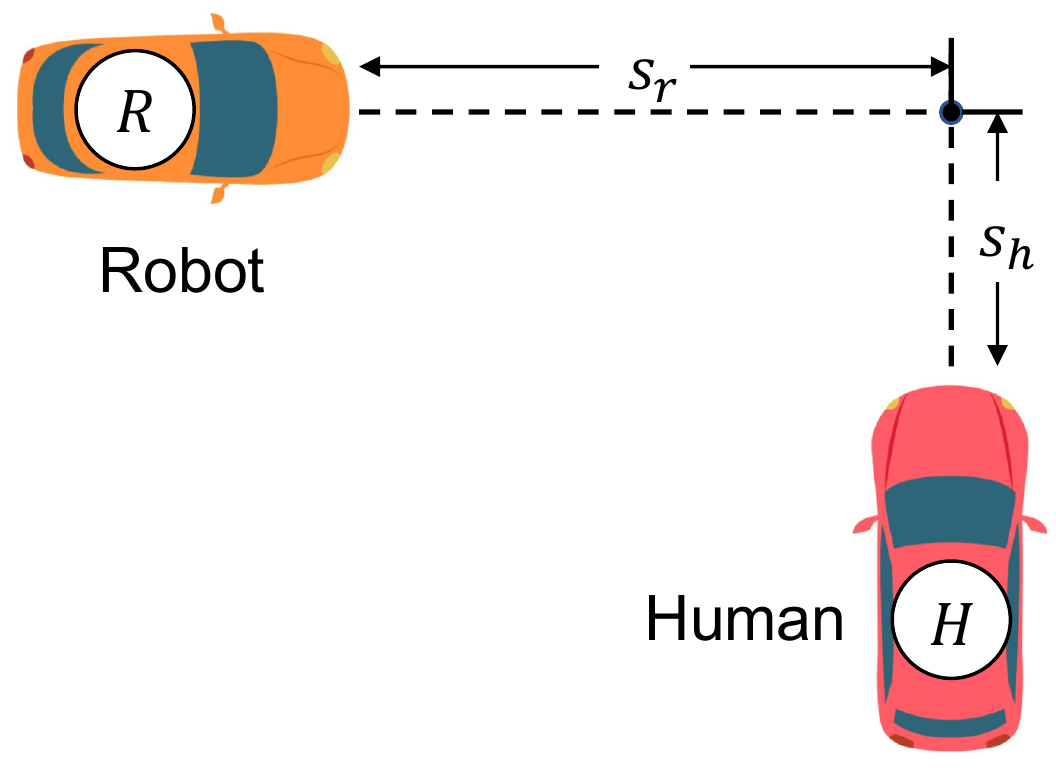}
    \caption{A motivating toy example, where a human car and a robot car are driving towards a collision point.}
    \label{fig:toy_example}
    \vspace{-0.1in}
\end{figure}

We begin our discussion by studying a motivating toy example to demonstrate the issue of using conditional inference for interactive prediction. We consider the example depicted in Fig. \ref{fig:toy_example}, where two cars are driving towards a collision point. One of them is controlled by a human driver, while the other is an autonomous robot car. As an analogy of the CBP task, the robot car can query the posterior distribution of the human car's trajectory conditioned on a planned trajectory of the robot car. The robot car can then evaluate the risk of multiple trajectories and select the optimal one to execute.

We model the human driver's behavior with the intelligent driver model (IDM) \cite{treiber2013traffic, treiber2000congested}. In the IDM model, each car has two states at each timestep, $s_{i,t}$ and $v_{i,t}$, where $s_{i,t}$ is the displacement relative to the collision point and $v_{i,t}$ is the absolute velocity. We denote the state vector by $\mathbf{x}_{i,t}=\left[s_{i,t}\ v_{i,t}\right]^\intercal$. Each car is assigned a target position to follow at each step, depending on which car has the right-of-way. If a car has the right-of-way, it is asked to follow a target substantially far away. Otherwise, if the other car has the right-of-way and has not passed the collision point, the target is set as the collision point. We determine which car has the right-of-way based on which car has smaller time headway at the current timestep. The time headway is defined as follows:
\begin{equation}
    T_{head, i, t} = \max\left(\frac{s_{i,t}}{v_{i,t}}, 0\right).
\end{equation}

Given a target position, the car dynamics is governed by the intelligent driver model as follows:
\begin{equation}
\begin{split}
    s_{i, t+1} &= s_{i,t} - \Delta t \cdot v_{i,t},  \\
    v_{i, t+1} &= v_{i, t} + \Delta t \cdot \omega_{i,t} \\
     + \Delta t \cdot &\left\{a\left[1 - \left(\frac{v_{i,t}}{v_0}\right)^\delta - \left(\frac{s^*(v_{i,t},\Delta v_{i,t})}{s_{i,t}-d_{i,t}}\right)^2\right]\right\},  
\end{split}
\end{equation}
where 
\begin{equation}
    \begin{split}
        \Delta v_{i, t} &= v_{i,t} - v_0,  \\
        s^*(v_{i,t},\Delta v_{i,t}) &= s_0 + \max\left(0, v_{i,t}T + \frac{v_{i,t}\Delta v_{i, t}}{2\sqrt{ab}} \right),  \\
        \omega_{i,t} &\sim \mathcal{N}\left(0, \sigma^2\right).
    \end{split}
\end{equation}
The term $d_{i,t}$ denotes the target position. It equals to zero if the target point collides with the collision point. Otherwise, a large negative value is assigned to $d_{i,t}$. The Gaussian noise $\omega_{i,t}$ is added to inject randomness. The remaining parameters are defined as in the standard IDM model. Readers may refer to \cite{treiber2013traffic} for detailed definitions. In our experiments, we set $v_0=10\mathrm{m/s}$, $T=2\mathrm{s}$, $s_0=4\mathrm{m}$, $\delta=4$, $a=1\mathrm{m/s^2}$, $b=1.5\mathrm{m/s^2}$, $\Delta t=0.2\mathrm{s}$, and $\sigma=4\mathrm{m/s^2}$. 

\begin{figure}[t]
    \centering
    \subfloat{{\includegraphics[width=5.1cm]{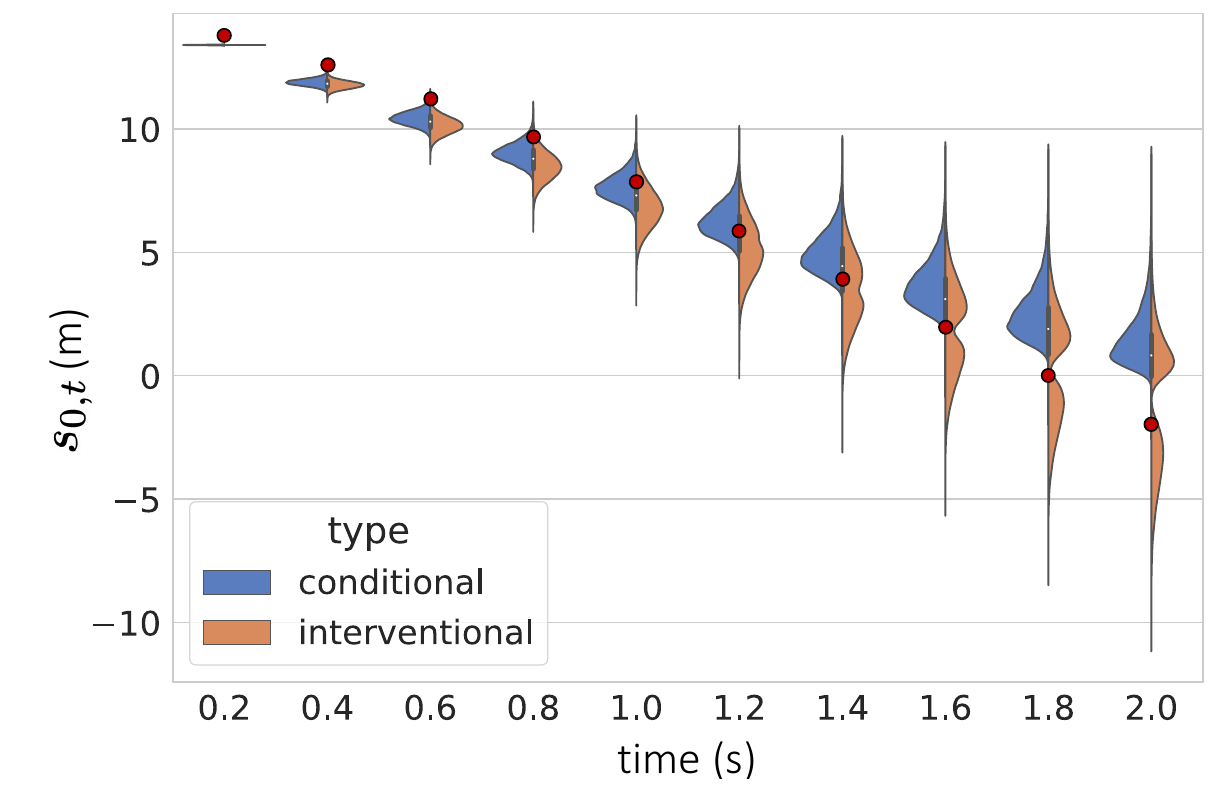} }}
    \qquad
    \subfloat{{\includegraphics[width=5cm]{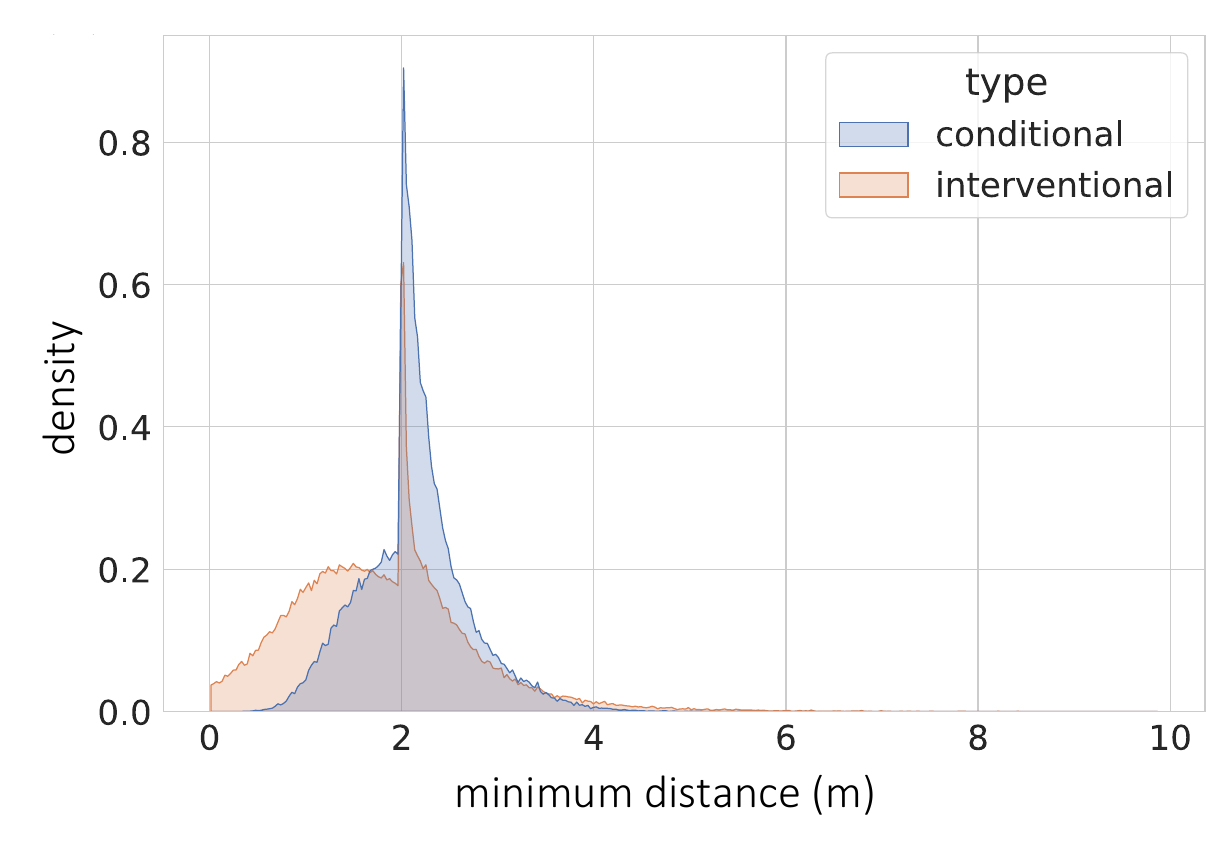} }}
    \caption{{\bf Top}: Histograms of $s_{h,t}$ for $t=1,\cdots,T_H$ under the conditional distribution and the distribution given the interventional action. The red dots denote the planned trajectory of the robot car; {\bf Bottom}: Normalized histograms of minimum distance between cars under these two distributions. The histograms are drawn with 10000 simulation trials.}
    \label{fig:toy_result}
    \vspace{-0.1in}
\end{figure}

The robot car predicts the human-driven car's trajectory based on its behavior when both cars are controlled by humans, i.e., governed by the IDM model. In the CBP task, we aim to approximate the distribution of the human-driven car's future trajectory conditioned on the initial states of the two cars and the future trajectory of the robot car, i.e., $p(\mathbf{x}_{h,1:T_H}|\mathbf{x}_{h,0},\mathbf{x}_{r,0},\mathbf{x}_{r,1:T_H})$, where $T_H$ denotes the number of timesteps. The robot car can query the conditional distribution with a planned trajectory $\mathbf{\hat{x}}_{r,1:T_H}$. In our experiment, we use likelihood weighting \cite{russel2010} to estimate the conditional distribution given an evidence set $\left\{\mathbf{\hat{x}}_{h,0},\mathbf{\hat{x}}_{r,0},\mathbf{\hat{x}}_{r,1:T_H}\right\}$. Meanwhile, we can approximate the actual distribution of $\mathbf{x}_{h,1:T_H}$ after executing $\mathbf{\hat{x}}_{r,1:T_H}$ via multiple simulation trials. In Fig.~\ref{fig:toy_result}(a), we compare the two distributions under the same initial conditions and query trajectory. We set $s_{h,0}=s_{r,0}=15\mathrm{m}$, $v_{h, 0}=8\mathrm{m/s}$, $v_{r, 0}=5\mathrm{m/s}$, and $T_H=10$. Since the robot car has smaller initial speed, it is more likely to yield to the human car. However, we let the robot car execute an aggressive maneuver, where the robot car accelerates with an acceleration of $5\mathrm{m/s^2}$ until reaching the speed of $10\mathrm{m/s}$. 

The conditional distribution implies that the human car always yields to the robot car. However, the human car may actually not yield to the robot car when the robot car executes the planned trajectory. Even if the human car eventually yields to the robot car, it starts decelerating much later than the conditional distribution suggests. If we evaluate the risk based on the conditional distribution, we may falsely conclude that the human car will always yield to the robot car, so that the robot car can safely pass the intersection at high speed, which leads to an overly aggressive and unsafe maneuver. It can be further verified by estimating the histograms of the minimum distance between the two cars under these two distributions (Fig. \ref{fig:toy_result}(b)). The minimum distance is biased under conditional inference. In particular, it falsely implies that the two cars never collide. 

\begin{figure}[t]
    \centering
    \includegraphics[width=3.2in]{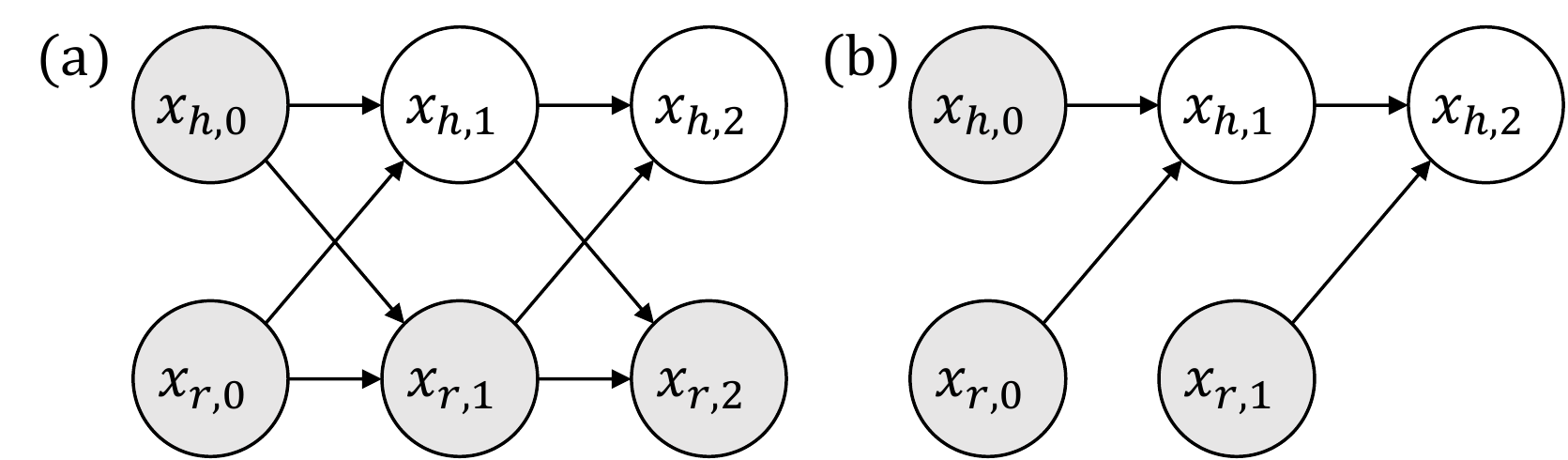}
    \caption{(a) The Bayesian network representing the conditional distribution $p(\mathbf{x}_{h,1:T_H}|\mathbf{\hat{x}}_{h,0},\mathbf{\hat{x}}_{r,0},\mathbf{\hat{x}}_{r,1:T_H})$; (b) The Bayesian network representing the distribution resulting from the intervention $do(\mathbf{x}_{r,1:T_H}=\mathbf{\hat{x}}_{r,1:T_H})$, denoted by $p(\mathbf{x}_{h,1:T_H}|\mathbf{\hat{x}}_{h,0},\mathbf{\hat{x}}_{r,0},do(\mathbf{\hat{x}}_{r,1:T_H}))$.}
    \label{fig:bayes_net}
    \vspace{-0.1in}
\end{figure}

The toy example demonstrates the discrepancy between the reality and the anticipation from conditional inference. Formally, the conditional distribution is governed by the Bayesian network in Fig. \ref{fig:bayes_net}(a), where the initial states and the query trajectory are treated as an \emph{observation}. However, the system is actually governed by the Bayesian network in Fig. \ref{fig:bayes_net}(b) when the robot executes $\mathbf{\hat{x}}_{r,1:T_H}$. The incoming edges of $\mathbf{x}_{r,i}$ are removed because the robot car follows a fixed trajectory regardless of the other car's reaction. If fact, if we treat the Bayesian network governing the system as a causal Bayesian network \cite{reason:Pearl09a}, then Fig. \ref{fig:bayes_net}(b) represents the distribution resulting from the interventional action $do(\mathbf{x}_{r,1:T_H}=\mathbf{\hat{x}}_{r,1:T_H})$, denoted by $p(\mathbf{x}_{h,1:T_H}|\mathbf{\hat{x}}_{h,0},\mathbf{\hat{x}}_{r,0},do(\mathbf{\hat{x}}_{r,1:T_H}))$. The difference between the two distributions, $p(\mathbf{x}_{h,1:T_H}|\mathbf{\hat{x}}_{h,0},\mathbf{\hat{x}}_{r,0},\mathbf{\hat{x}}_{r,1:T_H})$ and $p(\mathbf{x}_{h,1:T_H}|\mathbf{\hat{x}}_{h,0},\mathbf{\hat{x}}_{r,0},do(\mathbf{\hat{x}}_{r,1:T_H}))$, mirrors the difference between seeing and doing \cite{reason:Pearl09a}. By conditional inference, we aim to infer the distribution of $\mathbf{x}_{h,1:T_H}$ after \emph{observing} $\mathbf{\hat{x}}_{h,1:T_H}$. Intuitively speaking, it represents how the human driver will react if he knows the robot car will execute $\mathbf{\hat{x}}_{r,1:T_H}$ in advance. However, we should not inform the human driver of the robot car's future motion when evaluating the consequence of the action $do(\mathbf{x}_{r,1:T_H}=\mathbf{\hat{x}}_{r,1:T_H})$. It leads to overly confident anticipation on human's reaction on aggressive maneuvers, as demonstrated in our toy example. Instead, we should evaluate $\mathbf{\hat{x}}_{h,1:T_H}$ with a model approximating the distribution $p(\mathbf{x}_{h,1:T_H}|\mathbf{\hat{x}}_{h,0},\mathbf{\hat{x}}_{r,0},do(\mathbf{\hat{x}}_{r,1:T_H}))$, in other words, a model designed for the IBP task. 

\section{Quantifying the Impact of Conditional Inference on Real-World Datasets} \label{sec:shapley}
From the toy example, we have made it clear that conditional inference leads to biased prediction and potential safety hazard. We are then curious about: 1) how conditional inference may impact interactive prediction in real-world scenarios; and 2) how we can identify a CBP model with potential safety risk. Unlike the toy example, we do not have access to the ground-truth dynamics governing the interacting agents. Meanwhile, it is expensive and dangerous to estimate and compare the conditional and interventional distributions via real-world experiments. Instead, we are interested in a method purely based on offline datasets. 

Intuitively, the direct consequence of treating the query trajectory as observation is that the robot may anticipate the interacting agents to react to its future actions in advance. Therefore, we propose to detect and quantify the impact of conditional inference by looking into how much the future segment of the query trajectory contributes to the prediction at prior timesteps for a given CBP model. In particular, we adopt Shapley values as the evaluation tool. 

\subsection{Shapley Value in Explainable Deep Learning}
Originated in cooperative game theory, Shapley values have been widely used in deep learning to quantify feature attribution of black-box models \cite{lundberg2017unified}. Shapley values quantify the attribution of each dimension of an input $x=(x_1,\cdots,x_n)$ to a function describing the model behavior $f:\mathcal{X}_1\times\cdots\times \mathcal{X}_n\rightarrow\mathbb{R}$. The output of $f$ could be the direct output of the model. Alternatively, $f$ could also output an numerical value quantifying the performance or uncertainty of the model. Formally, one defines a set function $\nu:\mathbb{S} \rightarrow \mathbb{R}$ where $\mathbb{S}$ is the power set of $N:=\left\{1, 2, \cdots, n\right\}$, i.e., $\mathbb{S}=P(N)$. For a subset $S\in \mathbb{S}$, the output $\nu(S)$ corresponds to running the model on a modified version of the input $x$ for which features not in $S$ are dropped or replaced. For instance, we may replace the dropped features $x_{N\setminus S}$ with samples drawn from their marginal distribution in the dataset \cite{lundberg2017unified}, and then define:
\begin{equation} \label{eqn:v}
    \nu(S)=\mathbb{E}\left[f(x_S, X_{N\setminus S})\right].
\end{equation}
For each feature $x_i$, its Shapley value $\phi(x_i)$ is defined as:
\begin{equation}
    \phi(x_i) = \sum_{S\subseteq N\setminus \{i\}} \frac{1}{n \binom{n-1}{|S|}}\left(\nu \left(S\cup\{i\}\right) - \nu \left( S\right)\right),
\end{equation}
i.e., the difference in $\nu$ between including and not including the feature $x_i$ averaged over all subsets $S$. In the context of trajectory prediction, Shapley values have been adapted to quantify the usage of social cues of prediction models \cite{makansi2021you}. 

\begin{figure*}[t]
    \centering
    \includegraphics[width=4.2in]{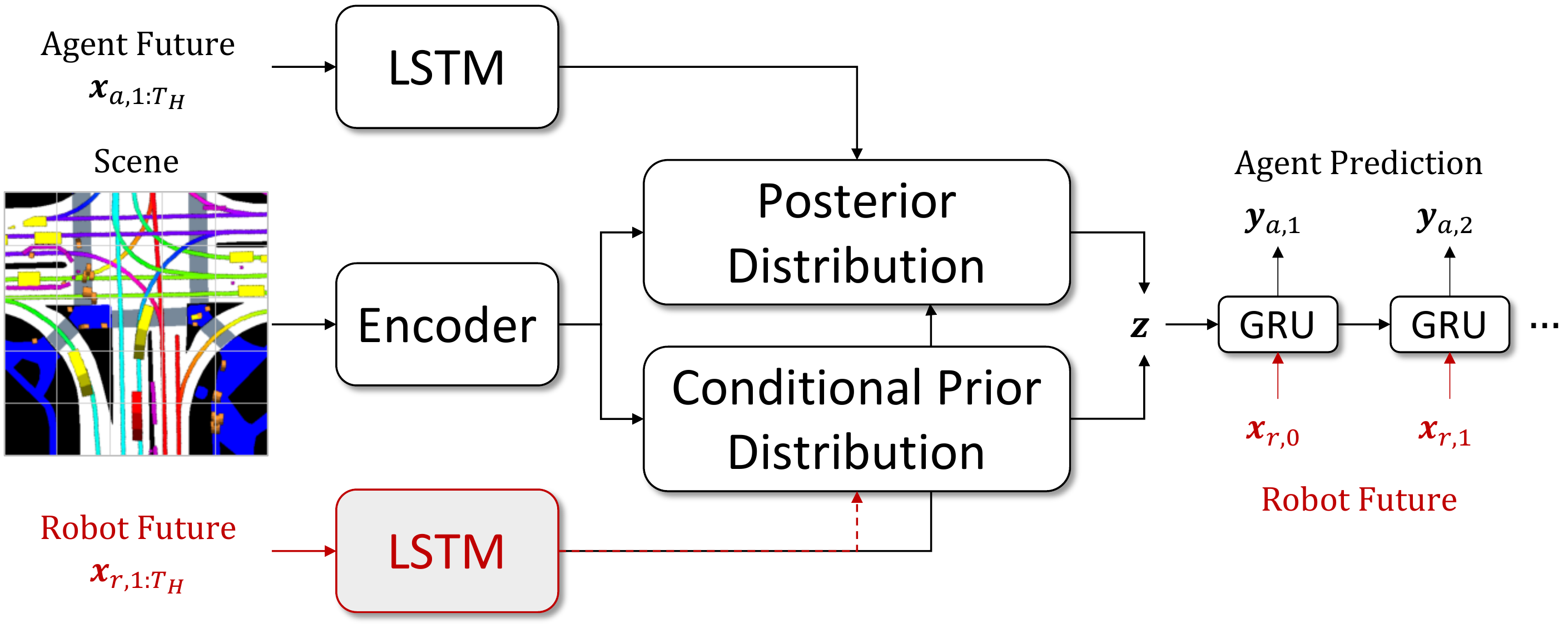}
    \caption{The scheme of conditional behavior prediction with Trajectron++.}
    \label{fig:trajectron++}
    \vspace{-0.1in}
\end{figure*}

\subsection{Shapley Value for Interventional Behavior Prediction} \label{sec:shapley-cbp}
In our case, we define the Shapley values of features based on their attribution to prediction performance. We adopt three evaluation metrics: 
\begin{itemize}
    \item \emph{Average Displacement Error (ADE)}: Mean $L_2$ distance between the ground-truth and predicted trajectories averaged over $K$ samples.
    \item \emph{Final Displacement Error (FDE)}: $L_2$ distance between the ground-truth and predicted final position averaged over $K$ samples.
    \item \emph{Kernel Density Estimate-based Negative Log Likelihood (KDE NLL)}: Mean NLL of the ground-truth trajectory under a distribution created by fitting a kernel density estimate on $K$ trajectory samples \cite{salzmann2020trajectron++}. 
\end{itemize}
It is worth noting that we do not follow the common practice to evaluate the minimum of distance metrics. As argued in \cite{makansi2021you}, computing the minimum leads to biased estimation. To minimize unnecessary bias in the computation of $\nu \left(S\cup\{i\}\right) - \nu \left( S\right)$, we choose a sufficiently large $K$ and only consider the average values of the distance metrics. 

The features of interest are essentially the states of the planned trajectory. The model is evaluated with the ground-truth future trajectory of the robot car, denoted by $\mathbf{x}_{r,1:T_H}$. Since additional future information is granted, we expect more accurate prediction in average after conditioning on the ground-truth robot future, which leads to non-negative Shapley values in general. The question remained is how to segment the robot future trajectory into features. Since the entire trajectory is typically treated as a sequence when encoded \cite{salzmann2020trajectron++}, perturbing the state at a single timestep may merely have minimal effect on the model output, as the encoder may manage to smooth out the perturbation. Also, treating the state at each timestep as a single feature leads to large $n$, which makes the computation of Shapley values expensive as $|P(N)|$ grows exponentially with $n$. Instead, we split ${\mathbf{x}}_{r,1:T_H}$ into $m$ segments, with each segment consists of states from multiple neighbouring timesteps:
\begin{equation}
    \mathbf{x}_{r,1:T_H} = \left[\mathbf{x}_{r,1:t_1},\ \mathbf{x}_{r,t_1:t_2},\ \cdots,\ \mathbf{x}_{r,t_{m-1}:t_{m}}\right].
\end{equation}
We are then interested in evaluating the attribution of future segments to the prediction at earlier timesteps. To this end, we compute the Shapley values $\phi(\mathbf{x}_{r,t_j:t_{j+1}})$ regarding the prediction over the first $t_1$ timesteps. If the model inherits the temporal independence of the interventional distribution, we expect a large value for $\phi(\mathbf{x}_{r,1:t_{1}})$ but nearly zero for the latter segments. If any of the Shapley values for $\mathbf{x}_{r, t_j:t_{j+1}}$ with $j\geqslant 1$ is instead significant, it indicates the model learns a distribution with notable discrepancy to the interventional distribution, which may cause safety issue if deployed on on-road autonomous vehicles. 

To compute the Shapley values, we need to define the set function in Eqn. (\ref{eqn:v}), which requires a marginal distribution of dropped features to define the expectation. Since we do not have access to the ground-truth marginal distribution, in practice, we can train a separate unconditioned prediction model as an approximation. It predicts the distribution of the robot future trajectory, i.e., $q(\mathbf{x}_{r, 1:T_H})$. Then we can approximate the expectation over the marginal distribution via sampling from $q(\mathbf{x}_{r, 1:T_H})$. Given $K$ samples of predicted robot trajectories, i.e., $\hat{\mathbf{x}}^k_{r,1:T_H}$ for $k=1,\cdots,K$, the set function is approximated as follows:
\begin{equation}
    \nu(S)\approx \frac{1}{K}\sum_{k=1}^{K} f(\tilde{\mathbf{x}}_{r,1:T_H}), 
\end{equation}
where $\tilde{\mathbf{x}}^k_{j:j+1} = \mathbf{1}(j\in{S})\mathbf{x}_{j:j+1} + \mathbf{1}(j\notin {S})\mathbf{\hat{x}}^k_{j:j+1}$. The resulting Shapley values imply the characteristics of the model when queried by a motion planner imitating human behavior. Alternatively, we may estimate $\nu(S)$ with the motion planner that will be deployed for a customized analysis. 

\section{Experiments} \label{sec:exp}
With the proposed toolkit, we now study the impact of conditional inference for a state-of-the-art model on a real-world prediction dataset.

\begin{figure}[t]
    \centering
    \subfloat[\centering $\phi^{\text{ADE}}_j$ for $j=1,2,3$.]{{\includegraphics[width=5.6cm]{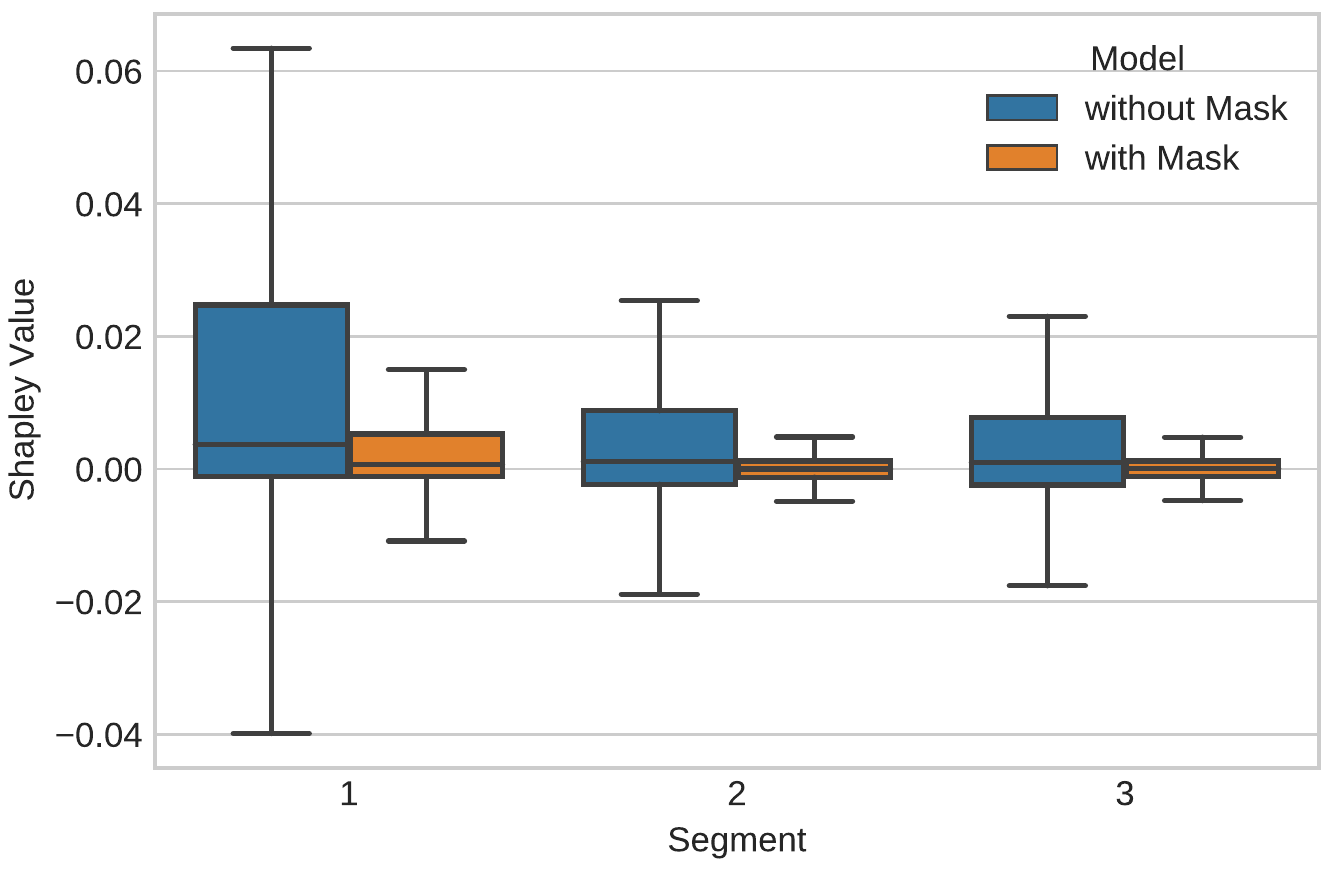}}}
    \
    \subfloat[\centering $\phi^{\text{FDE}}_j$ for $j=1,2,3$.]{{\includegraphics[width=5.6cm]{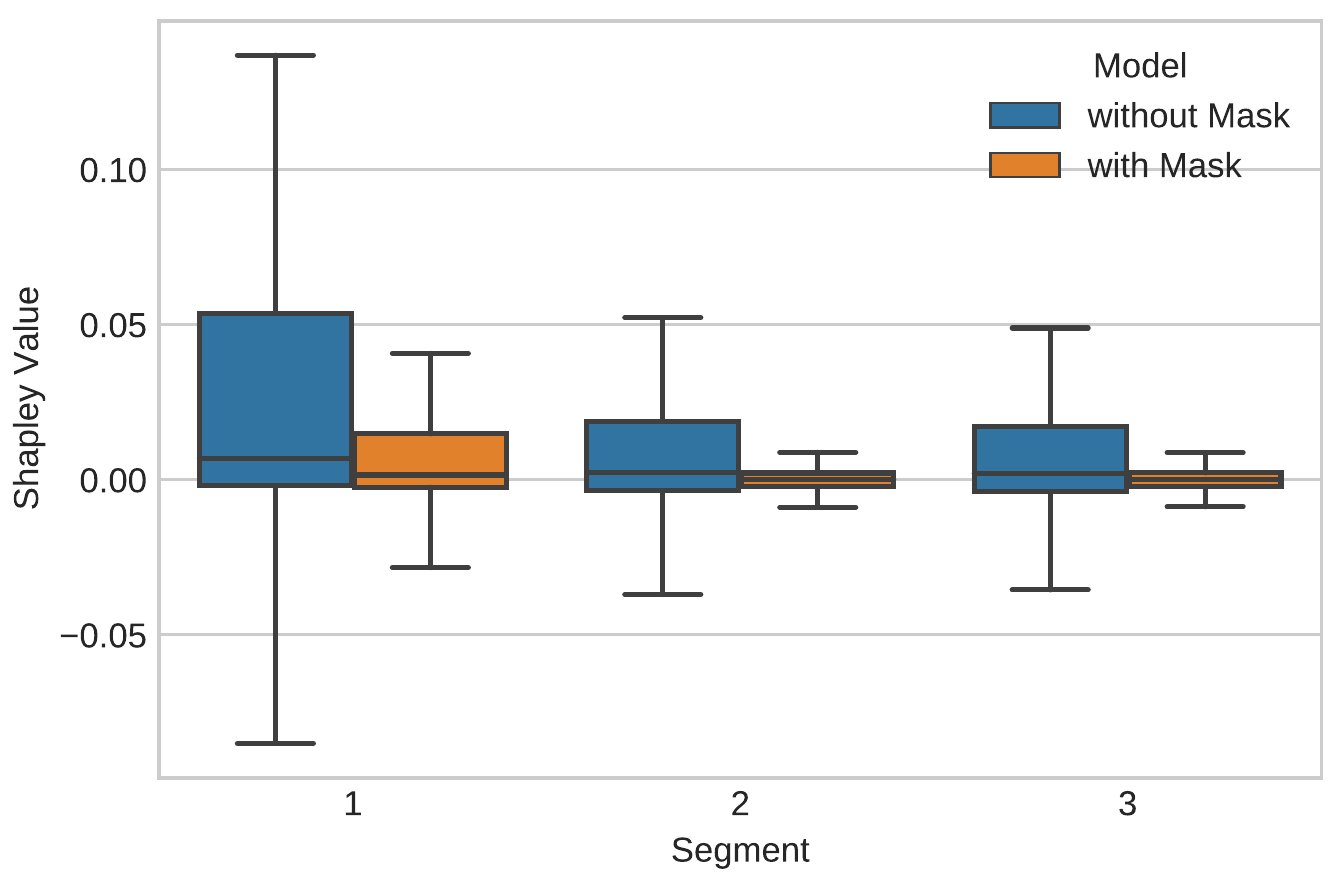}}}
    \
    \subfloat[\centering $\phi^{\text{KDE}}_j$ for $j=1,2,3$.]{{\includegraphics[width=5.6cm]{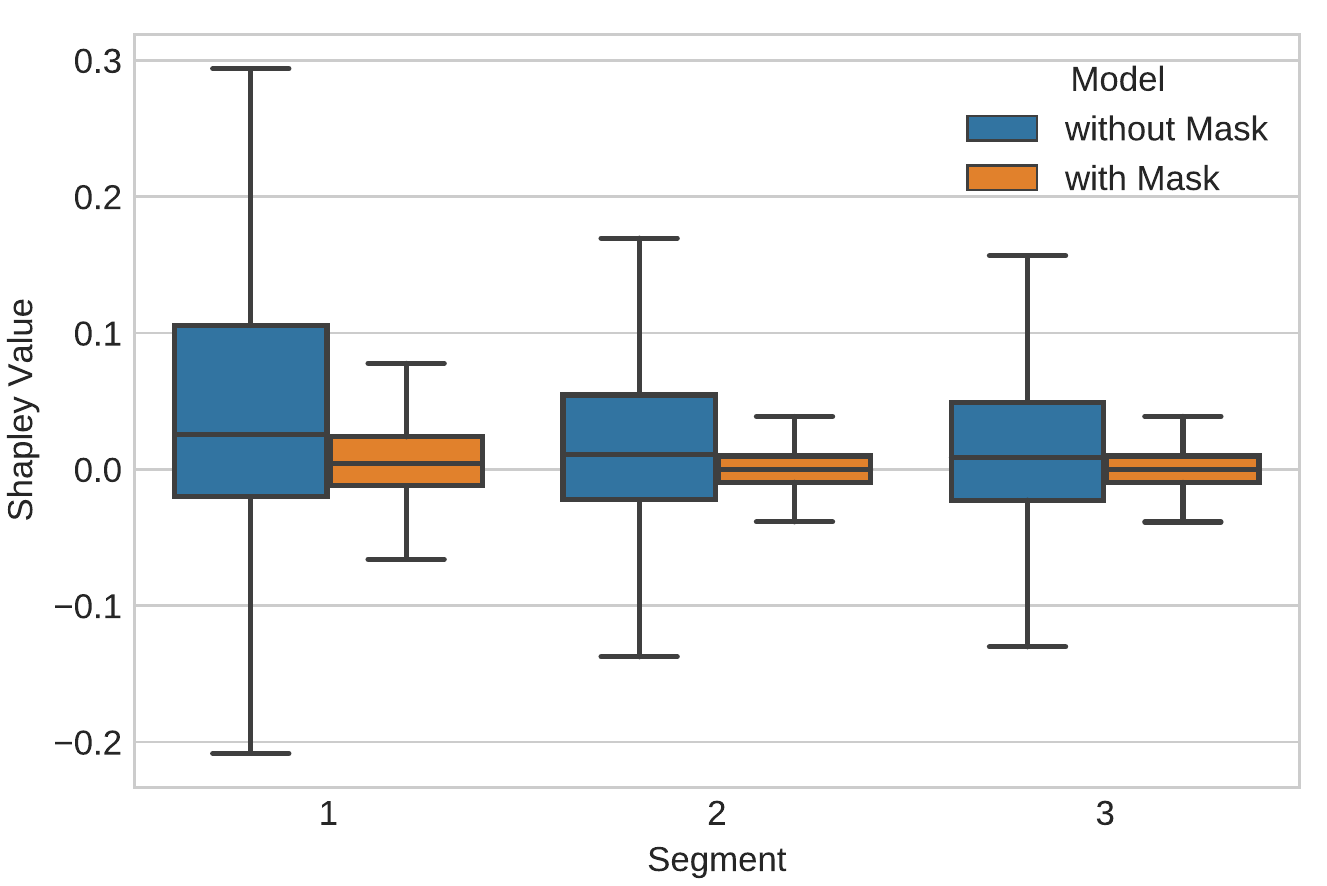}}}
    \caption{Box plots of Shapley values regarding different performance metrics. We compare the Shapley values of different segments of the robot future for the two models.}
    \label{fig:shapley_value} 
    \vspace{-0.1in}
\end{figure}

\subsection{Model and Dataset} 
We conduct our experiments with a state-of-the art trajectory prediction model, Trajectron++ \cite{salzmann2020trajectron++}, on the nuScenes dataset \cite{caesar2020nuscenes}. We choose Trajectron++ because it supports conditional trajectory prediction. More importantly, the authors showed that conditioning Trajectron++ on the future trajectory of the ego agent\textemdash referred to as \emph{robot future} in \cite{salzmann2020trajectron++}\textemdash indeed improved the model's prediction performance on the nuScenes dataset.

Trajectron++ leverages the Conditional Variational Autoencoder (CVAE) \cite{sohn2015learning} framework to explicitly model the multimodality in trajectory distribution. As shown in Fig. \ref{fig:trajectron++}, the robot future trajectory is fed into Trajectron++ through three channels: 1) feeding step-by-step into the corresponding Gated Recurrent Unit (GRU) cell \cite{cho2014learning} of the trajectory decoder; 2) feeding into the encoder modeling the posterior distribution of latent variables after encoded by a Long Short-Term Memory (LSTM) network \cite{hochreiter1997long}; 3) feeding into the encoder modeling the conditional prior distribution of latent variables after encoded by the same LSTM network. 

The first two channels are not problematic. The computational graph of the first channel is consistent with the causal Bayesian network after intervention as in Fig. \ref{fig:bayes_net}. For the second channel, the posterior distribution is only used during training. However, the last channel is potentially defective. During the inference stage, the latent variables are sampled from the conditional prior distribution. It takes the embedding generated by the LSTM network as input. The embedding fuses information along the entire planning horizon. Consequently, the model has access to the robot future states at latter timesteps when predicting the target agent's states at former timesteps. 

We are then curious about how much performance gain is attributed to this faulty shortcut. To this end, we use the Shapley values proposed in Sec. \ref{sec:shapley-cbp} to analyze the model behavior on the nuScenes dataset. The nuScenes benchmark sets a prediction horizon of 6 seconds. We then equally split the robot future trajectory into three segments. Afterwards, we compute the Shapley values to quantify their attribution to the prediction performance within the first 2 seconds in the future. In addition, we compare Trajectron++ with a variant of it, created by masking out the third channel (i.e., the dash line in Fig. \ref{fig:trajectron++}). By comparing the prediction performance of these two models, we can also have an idea on the effect of this shortcut on the model behavior. As mentioned in Sec.~\ref{sec:shapley-cbp}, we need to approximate the marginal distribution of the dropped features to compute the Shapley values. In our experiments, we simply train an unconditioned Trajectron++ model for this purpose. 

\begin{table}[b] 
\centering
\begin{threeparttable}
\caption{Shapley Values Comparison}\label{table:shapley}
\begin{tabular}{@{}cccc@{}} 
\toprule
Mask & $\phi^{\mathrm{ADE}}_1$ & $\phi^{\mathrm{ADE}}_2$ & $\phi^{\mathrm{ADE}}_3$ \\ \midrule
-    & $0.0148\pm 0.0839$      & $0.0049\pm 0.0444$      & $0.0044\pm 0.0376$      \\
\checkmark     & $0.0053\pm 0.0197$      & $0.0000\pm 0.0024$      & $0.0000\pm 0.0024$      \\ \toprule
Mask & $\phi^{\mathrm{FDE}}_1$ & $\phi^{\mathrm{FDE}}_2$ & $\phi^{\mathrm{FDE}}_3$ \\ \midrule
-    & $0.0332\pm 0.1569$      & $0.0117\pm 0.0829$      & $0.0109\pm 0.0716$      \\
\checkmark     & $0.0156\pm 0.0568$      & $0.0000\pm 0.0045$      & $0.0000\pm 0.0045$      \\ \toprule
Mask & $\phi^{\mathrm{KDE}}_1$ & $\phi^{\mathrm{KDE}}_2$ & $\phi^{\mathrm{KDE}}_3$ \\ \midrule
-    & $0.0636\pm 0.3365$      & $0.0192\pm 0.1725$      & $0.0179\pm 0.1676$      \\
\checkmark     & $0.0119\pm 0.0857$      & $0.0000\pm 0.0527$      & $0.0001\pm 0.0524$      \\ \bottomrule
\end{tabular}
\begin{tablenotes}
        \small 
        \item The results are presented in the format of $\mathrm{mean}\pm \mathrm{std}$.
\end{tablenotes}
\end{threeparttable}
\vspace{-0.1in}
\end{table}

\begin{table*}[t]
\centering
\begin{threeparttable}
\caption{Prediction Performance Comparison}\label{table:performance}
\begin{tabular}{@{}ccccc@{}}
\toprule
\multicolumn{2}{c}{{Ablation}} & \multirow{2}{*}{$min\mathrm{ADE}_{K=6}$} & \multirow{2}{*}{$min\mathrm{FDE}_{K=6}$} & \multirow{2}{*}{$\mathrm{KDE\ NLL}$} \\ \cmidrule(r){1-2}
{Robot} & {Mask} & & & \\ \midrule
- & - & $1.73\pm 2.32$ & $4.02\pm 5.62$ & $1.86\pm 3.23$ \\
\checkmark & - & $\mathbf{1.61\pm 2.44 \ (-6.94\%)}$ & $3.76\pm 5.99\ (-6.47\%)$ & $\mathbf{1.61\pm 3.73\ (-13.4\%)}$ \\
\checkmark & \checkmark & $\mathbf{1.61\pm 2.43 \ (-6.94\%)}$ & $\mathbf{3.72\pm 5.92\ (-7.46\%)}$ & $1.77\pm 3.43 \ (-4.84\%)$                      \\ \bottomrule
\end{tabular}
\begin{tablenotes}
        \small 
        \item The results are presented in the format of $\mathrm{mean}\pm \mathrm{std}$. The number in the parentheses indicates the percentage of improvement regarding the unconditioned model. 
\end{tablenotes}
\end{threeparttable}
\vspace{-0.1in}
\end{table*}

\subsection{Results}
We computed the Shapley values over the test set for the models with and without masking the channel feeding the robot future embedding into the conditional prior encoder. The results are summarized in Table \ref{table:shapley} and Fig. \ref{fig:shapley_value}, where we summarize the statistics of $\phi^\mathrm{ADE}_j$, $\phi^\mathrm{FDE}_j$, and $\phi^\mathrm{KDE}_j$ for $j=1,2,3$. The superscript denotes the corresponding performance metric. The subscript denotes the segment of the robot future trajectory. By masking the input channel, the model satisfies the temporal independence inherited in the causal Bayesian network after intervention. Therefore, it is as expected that the Shapley values are minimal for $j>1$. However, we can see from Fig. \ref{fig:shapley_value} that the values are not strictly zero for all the data samples, because of the randomness in model output. In contrast, the model without masking, i.e., the original Trajectron++ model, has significantly larger Shapley values for the latter two segments. More importantly, their magnitude is not negligible compared to the one of the first segment. It means the future states of the robot can falsely affect the model's prediction at earlier timesteps.

In summary, the Shapley values suggest that the CBP model is indeed biased and could potentially cause safety hazard after deployment. It is difficult though to precisely measure its consequence without online testing. Even during online testing, the effect of the biased prediction could only be observed in those highly interactive scenarios which make up a small proportion of real-world traffic scenarios. Therefore, we argue that a cheaper solution is to prevent the bias at the design stage. Instead of developing models for the CBP task, we should turn to the IBP task. The model should be carefully designed and implemented to follow an interventional distribution. Meanwhile, the proposed Shapley values should be used to monitor the model behavior. 

In particular, a prediction benchmark designed for the IBP task should include such a quantitative metric as a complement to the current prediction metrics. For instance, we may set constraints $\phi^\mathrm{FDE}_2\leqslant \epsilon$ and $\phi^\mathrm{FDE}_3\leqslant \epsilon$ for some small threshold value $\epsilon$. Only those models satisfying the constraints are qualified for performance comparison. Such constraints are crucial for prediction benchmark, because the models are evaluated as black boxes. It is expensive and time-consuming in general to check that leakage does not occur in the model design and implementation. Without the constraints, the performance comparison could be misleading and unfair. For instance, we compare the performance of the models with and without masking against the unconditioned model in Table \ref{table:performance}. While the masking only slightly affects the values of $min\mathrm{ADE}$ and $min\mathrm{FDE}$, the model without masking gains significant improvement in terms of $\mathrm{KDE\ NLL}$. Without the Shapley values, one may consider this model with the defective input channel a better prediction model.  

\section{Discussion} \label{sec:discuss}
\subsection{Training Prediction Model for IBP}
In our experiments, we demonstrated one practical way to design a prediction model for the IBP task. Same as a CBP model, the model takes the ego agent's future trajectory as input during training. However, we should ensure that the model architecture reserves the structure of the causal Bayesian network under intervention (e.g. Fig. \ref{fig:bayes_net}(b)). Alternatively, we may train a prediction model for the joint behavior of all the agents, including the ego agent and its surroundings. For online usage, we can then conduct intervention on this joint prediction model when given a planned trajectory. In fact, some prior works follow this scheme implicitly for conditional prediction \cite{schmerling2018multimodal, tang2019multiple, khandelwal2020if}. However, they mean to approximate the conditional distribution by enforcing the ego agent's action sequence, as it is intractable to conduct exact conditional inference on the joint prediction model. We are interested in formally comparing these two training schemes in the future. 

\subsection{Establishing Prediction Benchmark for IBP}
To compute the Shapley values in our experiments, we sample the ego agent's future trajectories from an unconditioned Trajectron++ model for our convenience. However, the sample distribution is sensitive to the training dataset. Also, if we want to establish a formal IBP benchmark, we cannot ensure a transparent and fair evaluation with a black-box sampling method for Shapley value computation. As a solution, we may evaluate the Shapley values with a set of plausible future trajectories generated by a model-based motion planner \cite{song2022learning}. In our future work, we will investigate it and develop IBP benchmarks on public datasets. 

Besides, we would like to emphasize that ensuring the temporal independence is only the basic requirement for a good IBP model. Since a planner may query an IBP model with an arbitrary planned trajectory, ideally the IBP model need to be accurate over the entire input space of planned trajectories. However, it is prohibitive in general to train such a perfect model with offline datasets. Instead, a practical solution is to equip the prediction model with a module detecting out-of-distribution inputs of planned trajectories \cite{filos2020can, sun2021complementing}, which can be utilized to prevent the planning module from exploiting the prediction model with those out-of-distribution inputs. Therefore, it is necessary to develop such a out-of-distribution detection module for an IBP model and include the evaluation on out-of-distribution inputs as a part of an IBP benchmark.

\section{Conclusion} \label{sec:conclusion}
In this work, we study the problem of conditional behavior prediction, which builds up the foundation for an interactive prediction and planning framework. We argue that it is risky for the planner to query a prediction model trained for the CBP task. Instead, we should treat the planned trajectory as an intervention and let the model learn the trajectory distribution under intervention, which we refer to as the IBP task. Moreover, to properly evaluate an IBP model with offline datasets, we propose a Shapley-value-based metric to verify if the prediction model satisfies the inherent temporal independence of an interventional distribution. We show that the proposed metric can effectively identify a CBP model violating the temporal independence. When establishing IBP benchmarks, we can then set constraints based on the proposed metric to ensure a fair evaluation of an IBP model. 



\bibliographystyle{ieeetr}
\bibliography{references}

\end{document}